\documentclass{article}

\usepackage{times}

\usepackage{graphicx}
\usepackage{subfig}

\usepackage{booktabs}
\usepackage{multirow}
\usepackage{color}

\usepackage[square,numbers]{natbib}

\usepackage[final]{nips_2018}

\usepackage{algorithm}
\usepackage{algorithmic}

\usepackage{amsmath}
\usepackage{bbm}
\usepackage{epstopdf}

\usepackage{wrapfig}
\usepackage{tikz}
\usepackage{pgfplots}
\usetikzlibrary{arrows,positioning,automata,bayesnet}
\usepgfplotslibrary{external} 

\usepackage[hyphens]{url}
\usepackage[hidelinks]{hyperref}
\hypersetup{breaklinks=true}
\usepackage{float}

\title{BCCNet: Bayesian classifier combination neural network}

\author{
  Olga Isupova$^{\star}$ \\
  \texttt{olga.isupova@eng.ox.ac.uk} \vspace{-1em}\\
   \And
  Yunpeng Li$^{\dagger}$ \\
  \texttt{yunpeng.li@surrey.ac.uk} \vspace{-1em}\\
  \And
  Danil Kuzin$^{\ddagger}$ \\
  \texttt{dkuzin1@sheffield.ac.uk} \vspace{-1em}\\
  \And
  Stephen J Roberts$^{\star}$ \\
  \texttt{sjrob@robots.ox.ac.uk} \vspace{-1em}\\
  \And Katherine Willis$^{\S}$\\
  \texttt{kathy.willis@zoo.ox.ac.uk}\\
  \And
  Steven Reece$^{\star}$ \\
  \texttt{reece@robots.ox.ac.uk} \\
  \And  \vspace{-2.5em}\\
  $^{\star}$Department of Engineering Science, University of Oxford, UK\\
  $^{\dagger}$Department of Computer Science,
  University of Surrey, UK \\
  $^{\ddagger}$Department of Automatic Control
  and Systems Engineering,
  University of Sheffield, UK \\
  $^{\S}$Department of Zoology,
  University of Oxford, UK\\
}

\begin{document}

\maketitle

\begin{abstract}
  Machine learning research for developing countries can demonstrate clear
  sustainable impact by delivering actionable and timely information to
  in-country government organisations (GOs) and NGOs in response to their critical
  information requirements. We co-create products with UK and in-country
  commercial, GO and NGO partners to ensure the machine learning algorithms
  address appropriate user needs whether for tactical decision making or
  evidence-based policy decisions.  In one particular case, we developed and
  deployed a novel algorithm, BCCNet, to quickly process large quantities of
  unstructured data to prevent and respond to natural disasters. Crowdsourcing
  provides an efficient mechanism to generate labels from unstructured data to
  prime machine learning algorithms for large scale data analysis. However,
  these labels are often imperfect with qualities varying among different
  citizen scientists, which prohibits their direct use with many
  state-of-the-art machine learning techniques.  We describe BCCNet, a framework
  that simultaneously aggregates biased and contradictory labels from the crowd
  and trains an automatic classifier to process new data. Our case studies,
  mosquito sound detection for malaria prevention and damage detection for
  disaster response, show the efficacy of our method in the challenging context
  of developing world applications.
\end{abstract}

\section{Introduction}

Wide area situation awareness or surveillance, for example, following a natural
disaster or preempting disease, benefits from rich, update-to-date yet
unstructured data, including post hurricane satellite imagery and malarial
mosquito audio signals.  A small amount of data labelled by hand through
crowdsourcing platforms like Zooniverse\footnote{\url{https://www.zooniverse.org}} can
be used to train machine learning algorithms, such as neural networks (NNs), to
label the rest of the data~\cite{gaunt2016}.  
However, the crowdsourced labels
can be noisy and inconsistent, posing enormous challenges for machine learning
algorithms to aggregate information and produce best decisions for policy makers
and rescue workers~\cite{poblet2017}.  The Bayesian classifier combination (BCC)
algorithm~\cite{simpson2013dynamic} resolves classifier bias and aggregates
labels taking classifier consistency into account.

\begin{wrapfigure}[22]{r}{0.3\textwidth}
\begin{center}
\includegraphics[width=0.3\textwidth]{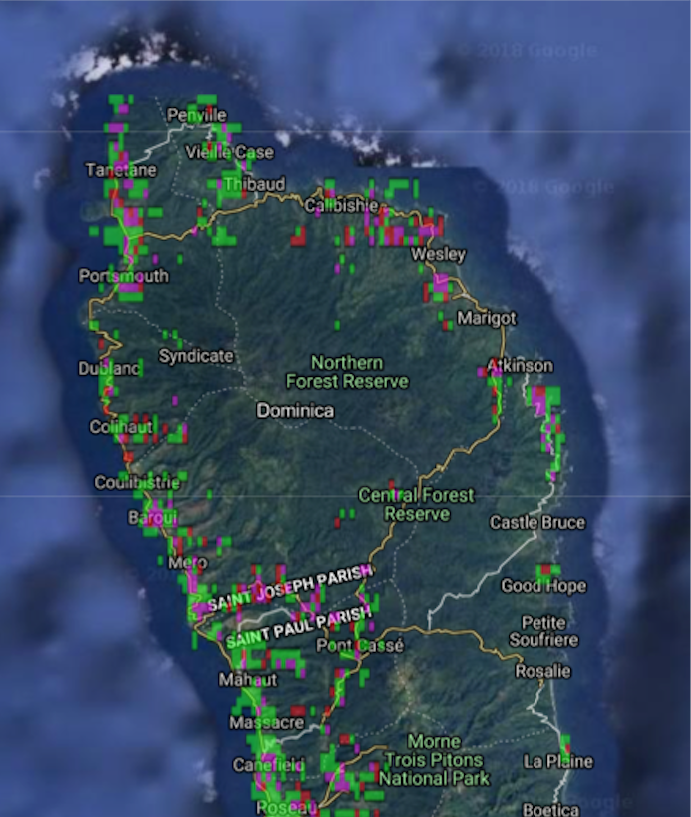}
\caption{Heatmap of building damage proportion in Northern Dominica after
  hurricane Maria in 2017: less than $20\%$ (green), $20\% \text{ to } 60\%$ (magenta),
  greater than $60\%$ (red).}
\label{fig:dominica_heatmap}
\end{center}
\end{wrapfigure} 
We propose an extension to BCC, the {\it Bayesian classifier combination neural
  network} (BCCNet), which incorporates a neural network object classifier.
BCCNet effectively trains the neural network object classifier using BCC bias
corrected crowd labels.  A novel hybrid variational Bayesian and maximum likelihood approach is developed to jointly learn
the neural network and BCC parameters.  We demonstrate the efficacy of the
approach on imbalanced data and biased crowd labels, scenarios common in real
applications.

Our algorithm has been developed and deployed in collaboration with Zooniverse
and Rescue Global\footnote{Rescue Global, Oxford machine
  learning and Zooniverse operational response team is collectively called the `Planetary
  Response Network'.}, a UK based not-for-profit, to
generate damage heatmaps for disaster responders by combining crowd labels of
satellite imagery immediately following Hurricanes Irma and
Maria (2017)~\cite{simpson2013dynamic,simpson2017bayesian, yore2017} (see
Figure~\ref{fig:dominica_heatmap}) and earlier versions following earthquakes in
Nepal (2015) and Ecuador (2016).  These heatmaps were passed to the UN, FEMA and
over 60 NGOs during the response phase of Irma and Maria in a timely
manner. This work has led to several research projects in disaster management
and environment protection in Africa, South East Asia and South America.
Our Zooniverse project on mosquito detection has crowdsourced labels from more
than 1200 citizen scientists on data collected in Thailand, Kenya, US and UK.

The rest of the paper is organised as follows: Section~\ref{sec:method} describes the BCCNet model. We present two case studies in Section~\ref{sec:experiment} and conclusions in Section~\ref{sec:conclusion}.


\section{The Bayesian Classifier Combination Neural Network Algorithm}
\label{sec:method}

BCCNet is a multi-class classifier that combines high dimensional data (e.g.,
images, audio signals) and noisy, potentially biased crowdsourced labels from a
set of imperfect base classifiers (e.g., crowd members).  It integrates a neural
network with the independent Bayesian classifier combination
algorithm~\cite{simpson2013dynamic}.

\begin{wrapfigure}[12]{l}{0.35\textwidth}
\begin{center}
\begin{tikzpicture}[x=1.7cm,y=1.8cm]

 \node[obs, inner sep = 0pt]    	(c_k)  {$c_i^{(k)}$}; %
 \node[latent, above=0.1 of c_k, xshift=38] (t) {$t_i$}
  		edge [->] (c_k); %
 \node[const, above=0.3 of t] (s) {$\mathbf{s}_i$}
  		edge [->] (t); %
  \node[const, above=0.3 of t, xshift=30] (theta) {$\boldsymbol\theta_{\text{NN}}$}
  		edge [->] (t); %
 \node[latent, above=0.1 of c_k, xshift=-38] (pi_k) {$\boldsymbol\pi^{(k)}$}
  		edge [->] (c_k); %
 \node[const, above=0.3 of pi_k]  	(alpha_k)		{$\boldsymbol\alpha_0^{(k)}$} 
  		edge [->] (pi_k); %
 {\tikzset{plate caption/.append style={below left=0pt and 0pt of #1.south east}}
 \plate {plate1} { %
     (c_k) %
     (t) %
     (s) %
  } {$i=[1,M]$}; %
  }%
  {\tikzset{plate caption/.append style={below right=0pt and 0pt of #1.south west}}
  \plate {plate2} { %
    (c_k) %
    (pi_k) %
    (alpha_k) %
  } {$k=[1, K]$}; %
  }%
  \draw[blue,thick,dashed] (1.05,0.8) ellipse (1.15cm and 1.05cm) +(0.15,-0.1) node[below right] {NN};
\end{tikzpicture}
\caption{Graphical model of BCCNet}
\label{fig:graph_model}
\end{center}
\end{wrapfigure}
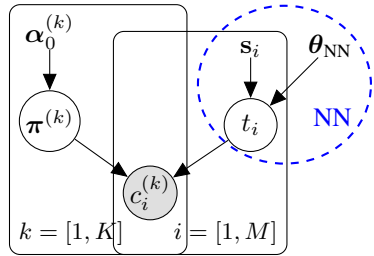
A neural network with parameters $\boldsymbol\theta_{\text{NN}}$ takes an
object~$\mathbf{s}_i$, e.g., an image patch of a satellite image, as input and
predicts a probability $p(t_i | \mathbf{s}_i, \boldsymbol\theta_{\text{NN}})$
that this object has class $t_i \in \{1, \ldots, J\}$, $\forall i \in \{1, \ldots, M\}$, where $M$ is the number of data points, and $J$ is the number of possible classes.

A label $c_i^{(k)} \in \{1, \ldots, L\}$ of a base classifier $k \in \{1, \ldots, K\}$ is drawn from the multinomial distribution depending on the true label for this data point:
\begin{equation}
\begin{split}
&c_{i}^{(k)} | \boldsymbol\pi^{(k)}, t_i \sim Mult(c_{i}^{(k)} ; \boldsymbol\pi_{t_i}^{(k)}) \\ 
&\forall i \in \{1,\ldots, M\}, k \in \{1, \ldots, K\},
\end{split}
\end{equation}
where $\boldsymbol\pi^{(k)}$ is a confusion matrix for the base classifier $k$, $\boldsymbol\pi_{t_i}^{(k)}$ is the $t_i$-th row of the confusion matrix~$\boldsymbol\pi^{(k)}$, $K$ is the total number of
base classifiers, $L$ is the number of values for the base classifiers' labels. 
Our approach tolerates the case when labels from the base classifiers are missing for some objects.

We impose a Dirichlet prior with hyperparameters $\boldsymbol\alpha_{0j}^{(k)}$ for rows of the confusion matrices:
\begin{equation}
\boldsymbol\pi_j^{(k)} | \boldsymbol\alpha_{0j}^{(k)} \sim Dir(\boldsymbol\pi_j^{(k)} ; \boldsymbol\alpha_{0j}^{(k)}), \forall j \in \{1, \ldots, J\}, k \in \{1, \ldots, K\}.
\end{equation}

The resulting graphical model is given in Figure~\ref{fig:graph_model}. BCCNet
inference is based on maximisation of the evidence lower bound (ELBO). The ELBO
is optimised using coordinate ascent over the NN parameters
$\boldsymbol\theta_{\text{NN}}$ and the posterior approximating distributions
for object class labels $t_i$ and confusion matrices $\boldsymbol\pi^{(k)}$
for the base classifiers. The NN parameters are updated via stochastic gradient
ascent and the posterior approximating distribution is found using the
variational mean-field approach. We iterate between one full pass of the data for the NN parameter update and one iteration for the approximating distribution update. We refer to this algorithm as {\it VB}.

\section{Experiments and results}
\label{sec:experiment}

We evaluated our approach on two real case studies, response after a natural
disaster and malaria prevention, and compared the proposed algorithm (VB) with
two baselines: \emph{i)} the EM-algorithm~\cite{albarqouni2016aggnet} (EM)
extended to our BCCNet model from Section~\ref{sec:method}, and \emph{ii)} the
neural network with an added crowd layer that models the confusion
matrices
~\cite{rodrigues2018deep} (CL). The base neural network for all methods was
LeNet-5~\cite{lecun1998gradient} with the Adam
optimiser~\cite{kingma2015adam}. The learning rate was chosen by grid search on
validation datasets. We also used validation datasets for early stopping. The
results are obtained from trained neural networks on held-out test datasets over
$30$ Monte Carlo runs with random initialisation.

\subsection{Case study 1: damage detection in satellite imagery for disaster response}
We analysed crowdsourced labels of damage from Digital
Globe\footnote{\url{https://www.digitalglobe.com}} high
resolution (30cm) optical satellite imagery of Dominica before and after
Hurricane Maria in 2017. Crowd members were presented with a subset of satellite
sub-images after the hurricane and asked, amongst other tasks, to draw bounding
boxes around all buildings in their sub-images and also mark building damage.

We extracted image patches from both `before' and `after' imagery corresponding to
the bounding boxes as input for a neural network. Image patches were
resized as $28 \times 28$ (the size of an average bounding box). Before and
after image patches formed different channels of the NN input layer. We also
extracted corresponding labels from the crowd as: ``background'', ``undamaged
building'', and ``damaged building''. We thus obtain a dataset with $M=32, 932$
objects labelled by $K=13$ volunteers (each object is labelled on average by $6$ volunteers). This dataset is challenging because of
the high discrepancy between different crowd members' answers: $38\%$ of the
objects were assigned to different classes by the crowd members (for comparison
in the second case study, below, the data had only $20\%$ of such objects).

The data lacked ground truth labels to validate the algorithms so we defined
ground truth as the crowd consensus output inferred using
BCC ~\cite{simpson2013dynamic} when the whole dataset was processed. We then
divided the dataset in the ratio $70-10-20\%$ into training, validation and test
datasets for evaluation of the algorithms.  The classification accuracy is given
in Figure~\ref{fig:catapult_results}. One can notice that the crowd layer network has the lowest accuracy. The VB algorithm for BCCNet provides not only the highest accuracy but also the most stable results among different Monte Carlo runs consistently for all three classes.

\subsection{Case study 2: mosquito detection in audio for malaria prevention}

The HumBug project\footnote{\url{http://humbug.ac.uk}} aims to detect malaria-vectoring
mosquitoes through their flight tones~\cite{li2017}.  A malaria epidemic can
occur a few weeks after initial impact of the disease and it is crucial to
monitor malaria vectors (i.e. \textit{Anopheles} species) and respond in the
early stages~\cite{waring2005}. As an initial step, we have launched a
crowdsourcing project on the Zooniverse
platform\footnote{\url{https://www.zooniverse.org/projects/yli/humbug}} to label
2-second length audio clips as containing ``mosquito sound'' or ``no mosquito
sound''. The project has attracted $1, 246$ volunteers up to date who have
labelled $55, 590$ audio clips from laboratory recordings collected in UK, US
and Kenya and field recordings from Thailand.  However, the crowd label matrix
$\mathbf{c}$ is still very sparse, $99.8\%$ of the matrix values are missing, so
we chose data clips that were labelled by at least $2$ volunteers as our
training dataset to ensure that our objects were assigned a class with some
confidence.  Consequently, we had $M = 22, 186$ and $K = 1, 128$ in this case.
We used a subset of laboratory recordings with labels provided by the research
team of the Humbug project as ground truth labels for test and validation datasets
with $M_{\text{test}} = 6, 651$ samples for testing and
$M_{\text{val}} = 3, 326$ samples for validation.

The neural network input comprised $20 \times 26$ sound `images' constructed
from audio clips where $26$ is the dimension of the mel-spectrum and $20$ is the
number of windows we used to divide each of the 2-second long audio clips.
Mosquito detection audio clips are naturally heavily imbalanced with most of
clips containing no mosquito sounds. According to the majority voted labels in
the training data there are only $21\%$ of clips containing mosquito sounds. In
these settings, the crowd layer neural network always predicts ``no
mosquito''. Therefore, for the CL algorithm we balanced the training dataset based on
majority voted labels. Both EM and VB algorithms for BCCNet are able to train
appropriate networks on the raw data.

\begin{figure}[!t]
\centering
\subfloat[]{
\includegraphics[scale=0.33]{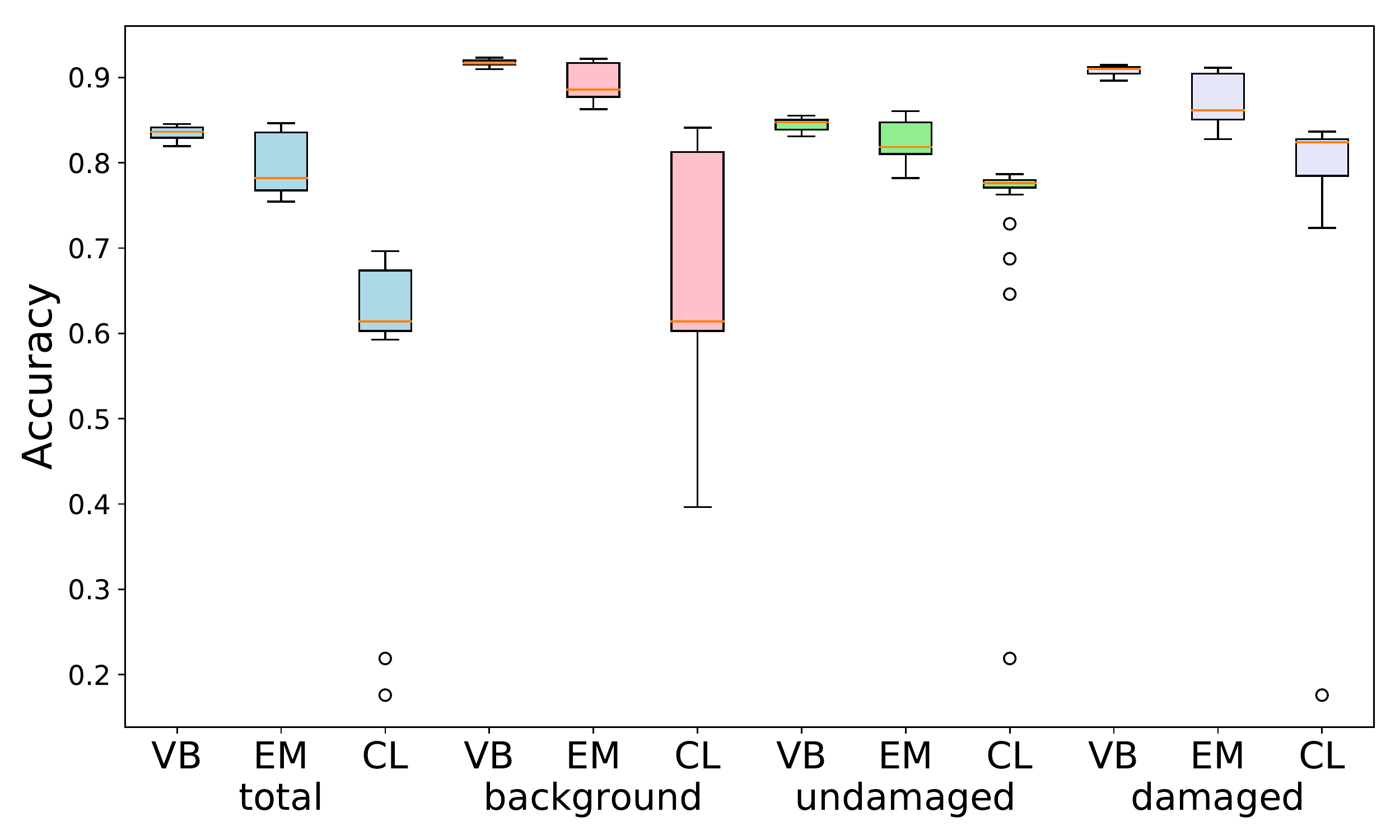}
\label{fig:catapult_results}}
\hfil
\subfloat[]{
\includegraphics[scale=0.33]{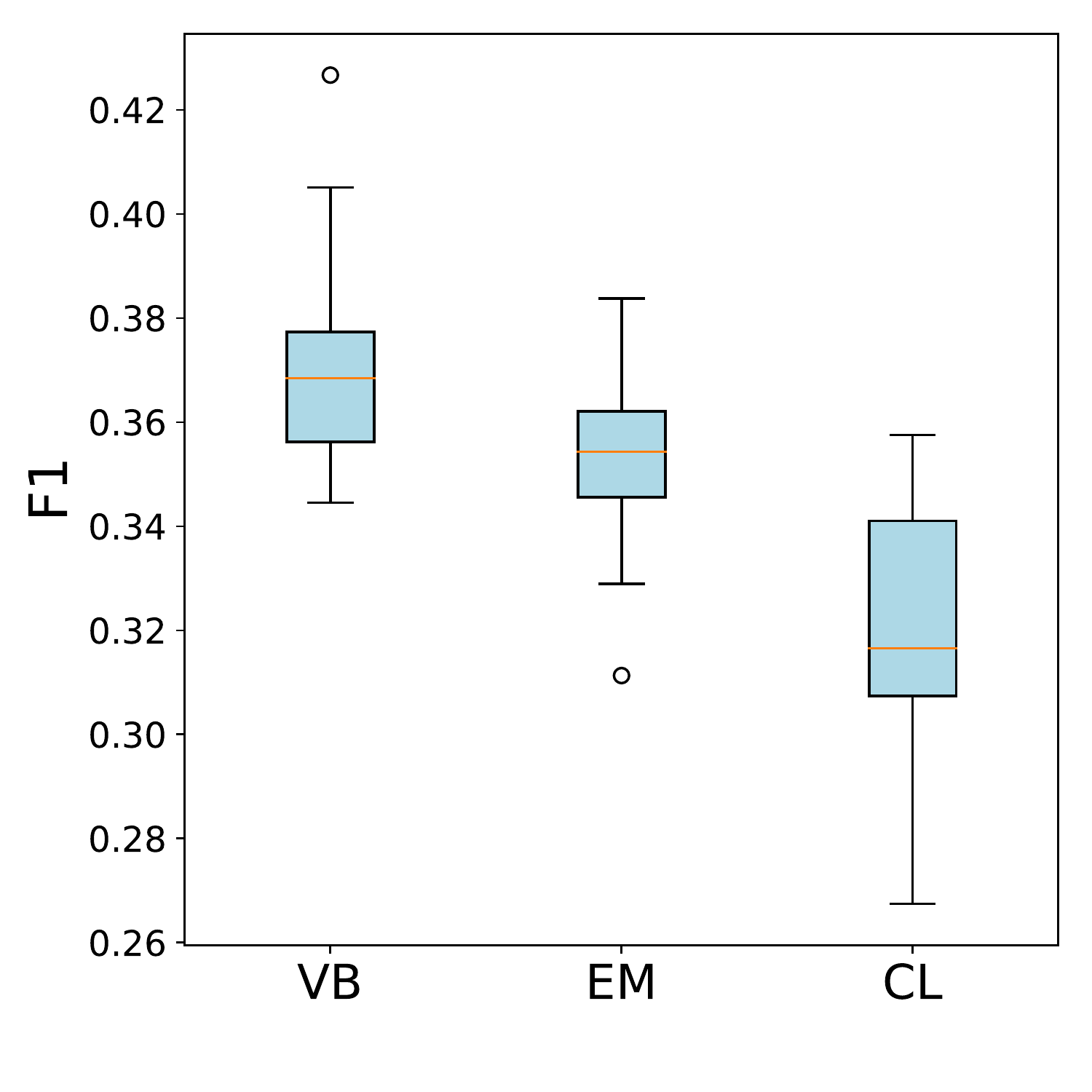}
\label{fig:humbug_results}}
\caption{Performance results. \protect\subref{fig:catapult_results} box plots for accuracy on the damage detection data: for all classes (blue), for the ``background'' class (red), for the ``undamaged
building'' class (green), and for the ``damaged building'' class (lavender). \protect\subref{fig:humbug_results} box plots for F1 measure on the mosquito detection data. }
\label{fig:results}
\end{figure}

Figure~\ref{fig:humbug_results} provides box plots of F1 measure for the
mosquito sound class. We used the F1 measure in this case as the data is 
highly imbalanced.  The crowd layer neural network has the lowest median
accuracy and the highest variance among different Monte Carlo runs. The
EM-algorithm for BCCNet provides more stable and more accurate results in
comparison to the crowd layer network. The proposed VB-algorithm for BCCNet also
gives stable results and additionally it has the highest median F1 measure amongst
the competitors.

\section{Conclusions}
\label{sec:conclusion}

We present BCCNet, an approach to jointly aggregate noisy crowdsourced labels
and train a neural network to process new data. This approach can be rapidly
deployed as a solution to challenging problems in the developing world that lack
labelled data. We demonstrate that BCCNet is stable, able to work with
imbalanced data and contradictory crowd labels. Ongoing operational engagement
with disaster responders shows that this technology delivers sustainable impact
by providing actionable and timely information to end users.

\subsubsection*{Acknowledgments}
This work is part-funded by a Google Impact Challenge award, by a grant
from the Alan Turing Institute's Data Centric Engineering programme and also
through the UK Space Agency's International Partnerships Programme.  The authors
would like to thank Digital Globe, Planet, ESA and the Satellite Applications
Catapult for ongoing satellite data provision; Dr. Marianne Sinka at the
University of Oxford, UK, Paul I. Howell at the Centers for Disease Control and
Prevention (CDC), BEI Resources in Atlanta, USA, Dustin Miller in CDC
Foundation, Centers for Disease Control and Prevention in Atlanta, Dr. Sheila
Ogoma, US Army Military Research Unit, Kisumu, Kenya (USAMRU-K), and
Dr. Theeraphap Chareonviriyaphap, Kasersart University, Thailand for their
collaborations on data collection and system deployment.

\small
\bibliography{Biblist}
\bibliographystyle{unsrtnat}

\end{document}